\documentclass[10pt,twocolumn,letterpaper]{article}

\usepackage{cvpr}
\usepackage{times}
\usepackage{epsfig}
\usepackage{graphicx}
\usepackage{amsmath}
\usepackage{amssymb}
\usepackage{comment}
\usepackage{multirow}
\usepackage{xcolor}
\usepackage{caption} 
\usepackage[pagebackref=true,breaklinks=true,letterpaper=true,colorlinks,bookmarks=false]{hyperref}

\usepackage[accsupp]{axessibility}  % Improves PDF readability for those with disabilities.

\cvprfinalcopy % *** Uncomment this line for the final submission

\definecolor{lightgray}{RGB}{220,220,220}
\definecolor{darkblue}{RGB}{0,0,127}
\definecolor{darkgreen}{RGB}{0,127,0}
\definecolor{darkred}{RGB}{200,0,0}

\ifcvprfinal\fi

\usepackage[affil-it]{authblk}

\usepackage{enumitem}

\title{The 6th AI City Challenge}

\begin{document}

\pagenumbering{gobble}

\author{
%\vspace{-0.7cm}
Milind Naphade$^1$ \hspace{0.9cm}
Shuo Wang$^1$ \hspace{0.9cm}
David C. Anastasiu$^2$ \hspace{0.9cm}
Zheng Tang$^1$ \\
Ming-Ching Chang$^3$ \hspace{0.9cm}
Yue Yao$^4$ \hspace{0.9cm}
Liang Zheng$^4$ \hspace{0.9cm}
Mohammed Shaiqur Rahman$^6$ \\
Archana Venkatachalapathy$^6$ \hspace{0.9cm}
Anuj Sharma$^6$ \hspace{0.9cm}
Qi Feng$^7$ \hspace{0.9cm}
Vitaly Ablavsky$^8$ \\
Stan Sclaroff$^7$ \hspace{0.9cm}
Pranamesh Chakraborty$^5$ \hspace{0.9cm}
Alice Li$^1$ \hspace{0.9cm}
Shangru Li$^1$ \\
Rama Chellappa$^9$
} 
\affil{ %\small
$^1$ NVIDIA Corporation, CA, USA \hspace{0.8cm} 
$^2$ Santa Clara University, CA, USA \\ 
$^3$ University at Albany, SUNY, NY, USA \hspace{0.8cm} 
$^4$ Australian National University, Australia \\
$^5$ Indian Institute of Technology Kanpur, India \hspace{0.8cm}
$^6$ Iowa State University, IA, USA \\
$^7$ Boston University, MA, USA \hspace{0.8cm}
$^8$ University of Washington, WA, USA \\
$^9$ Johns Hopkins University, MD, USA
}

\maketitle

%%%%%%%%%%%%%%%%%%%%%%%%%%%%%%%%%%%%%%%%%%%%%%%%%%%%%%%%%%
\begin{abstract}
The 6th edition of the AI City Challenge specifically focuses on problems in two domains where there is tremendous unlocked potential at the intersection of computer vision and artificial intelligence:  Intelligent Traffic Systems (ITS), and brick and mortar retail businesses. The four challenge tracks of the 2022 AI City Challenge received participation requests from 254 teams across 27 countries. Track 1 addressed city-scale multi-target multi-camera (MTMC) vehicle tracking. Track 2 addressed natural-language-based vehicle track retrieval. Track 3 was a brand new track for naturalistic driving analysis, where the data were captured by several cameras mounted inside the vehicle focusing on driver safety, and the task was to classify driver actions. Track 4 was another new track aiming to achieve retail store automated checkout using only a single view camera. We released two leader boards for submissions based on different methods, including a {\em public} leader board for the contest, where no use of external data is allowed, and a {\em general} leader board for all submitted results. The top performance of participating teams established strong baselines and even outperformed the state-of-the-art in the proposed challenge tracks. 
\end{abstract}

%%%%%%%%%%%%%%%%%%%%%%%%%%%%%%%%%%%%%%%%%%%%%%%%%%%%%%%%%%
%%%%%%%%%%%%%%%%%%%%%%%%%%%%%%%%%%%%%%%%%%%%%%%%%%%%%%%%%%
\section{Introduction}

AI has the potential to impact how we work, live and play. In the sixth edition of the AI City challenge we focus on challenge tasks that help make our experiences friction-less. While moving around cities, this means having AI improve our traffic systems to avoid congestion and ensuring driver safety. On the other hand when we are shopping in retail stores, making that experience friction-less translates into the ability to seamlessly walk in and out of a store with the least amount of time spent at the retail checkout. The common thread in making our experiences friction-lness across these two totally different environments boils down to the diverse uses of AI to extract actionable insights from a variety of sensors. We solicited original contributions in these and related areas where computer vision, natural language processing, and deep learning have shown promise in achieving large-scale practical deployment. To accelerate the research and development of techniques for these challenge tasks, we have created two new datasets. A brand new track and dataset around naturalistic driving behavior analysis was added, where the data were captured by several cameras mounted inside the vehicle focusing on driver safety, and the task was to classify driver actions. We also added a new track evaluating the accuracy of retail store automated checkout using only computer vision sensors. To this end, we released labeled data for various views of typical retail store goods with the evaluation focused on accurately recognizing and counting the number of such objects at checkout while accounting for clutter, and inter-object visual similarity and occlusions.

The four tracks of the AI City Challenge 2022 are summarized as follows:

\begin{itemize}[leftmargin=12pt] %\itemsep -.2em

\item \textbf{City-scale multi-target multi-camera (MTMC) vehicle tracking:} Participating teams were given video sequences captured at multiple intersections in a mid-sized city. The task is to track vehicles that pass through the field of views of multiple sensors. The evaluation is conducted on the \textit{CityFlowV2} dataset, including $313,931$ bounding boxes for $880$ distinct annotated vehicle identities.

\item \textbf{Tracked-vehicle retrieval by natural language descriptions:} 
This task offers natural language (NL) descriptions for tracked-vehicle targets in videos.  Participant teams are given videos with tracked-vehicle targets and NL queries to perform retrieval of the targets for each query.  The evaluation is conducted on 184 held-out queries and tracked-vehicles using the standard retrieval metric of Mean Reciprocal Rank (MRR).

\item \textbf{Naturalistic driving action recognition:}
In this track, teams are required to classify 18 different distracted behavior activities performed by the driver, such as texting, phone call, yawning, {\em etc.} The synthetic distracted driving (\textit{SynDD1}~\cite{https://doi.org/10.48550/arxiv.2204.08096}) dataset used in this track was collected using three cameras located inside a stationary vehicle. The training set consists of 30 videos and manually annotated files for each video stating the start and end time for every 18 tasks. The test set also consists of 30 videos but without any annotation file. Each video is in 1920×1080 resolution and around 10 minutes long.

\item \textbf{Multi-class product recognition \& counting for automated retail checkout:} The aim is to identify and count products as they move along a retail checkout lane. For example, given a checkout snapshot/video, teams need to identify and count all products, which may be very similar to each other or occluded by hands. One distinction about this track is that this track provides only synthetic data for model training. The provided synthetic training data come with various environmental conditions, while the real-world validation and test data are provided in the convenience of model distributed on real scenarios.

\end{itemize}

Consistent with the trend from past AI City Challenges, there was significant interest and participation in this year's Challenge. Since the challenge tracks were released in late February, we have received participation requests from 254 teams, which include 646 individual researchers from 181 recognized institutions across 27 countries. There were 194, 141, 150, and 125 participating teams in the 4 challenge tracks, respectively. The number of teams signing up for the evaluation system grew from 137 to 147 this year, where 119 of them submitted results to the leader boards. The four challenge tracks received 58, 24, 41, and 26 submissions, respectively. 

The paper summarizes the preparation and results of the 6th AI City Challenge. In the following sections, we describe the challenge setup ($\S$~\ref{sec:challenge:setup}), challenge data preparation ($\S$~\ref{sec:dataset}), evaluation methodology ($\S$~\ref{sec:eval}), analysis of submitted results ($\S$~\ref{sec:results}), and a brief discussion of insights and future trends ($\S$~\ref{sec:conclusion}).

%%%%%%%%%%%%%%%%%%%%%%%%%%%%%%%%%%%%%%%%%%%%%%%%%%%%%%%%%%
%%%%%%%%%%%%%%%%%%%%%%%%%%%%%%%%%%%%%%%%%%%%%%%%%%%%%%%%%%
\section{Challenge Setup}
\label{sec:challenge:setup}

The 6th AI City Challenge was set up in a similar format resembling the previous years. The training and test sets were released to the participants on February 27, 2022. All challenge track submissions were due on April 13, 2022. All the competitors for prizes were requested to release their code for validation. A new requirement for this year is that teams need to make their code repositories public, because we expect the winners to properly contribute to the community and the body of knowledge. The results on the leader boards have to be reproducible with no use of any external data.

\textbf{Track 1: City-Scale MTMC Vehicle Tracking.} Participating teams track vehicles across multiple cameras both at a single intersection and across multiple intersections spread out across a city. This helps traffic engineers understand journey times along entire corridors. The team with the highest accuracy in tracking vehicles that appear in multiple cameras is declared the winner of this track. In the event that multiple teams perform equally well in this track, the algorithm needing the least amount of manual supervision is chosen as the winner.

\textbf{Track 2: Tracked-Vehicle Retrieval by Natural Language Descriptions.}
In this challenge track, teams were asked to perform tracked-vehicle retrieval given single-view videos with tracked-vehicles and corresponding NL descriptions of the targets. Following the same evaluation setup used in the previous year, the performance of the retrieval task was evaluated using MRR. The NL based vehicle retrieval task offered unique challenges. In particular, different from prior content-based image retrieval systems~\cite{guo2018dialog,hu2016natural, mao2016generation}, retrieval models for this task needed to consider both the relation contexts between vehicle tracks and the motion within each track.

\begin{figure}[t]
\centering
\includegraphics[width=0.47\textwidth]{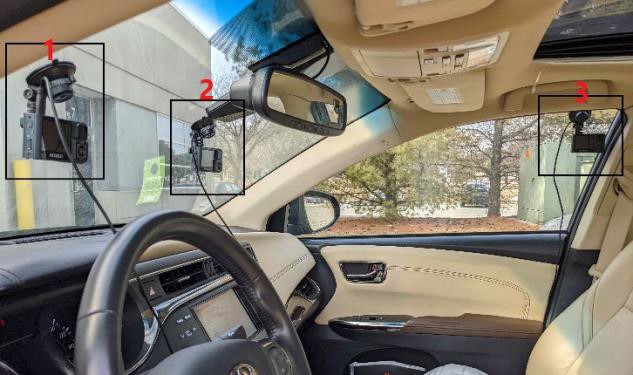}
\caption{Camera mounting setup for the three views listed in Table~\ref{tab:driver:view}.}
\label{fig:track3:camera}
\end{figure}

\textbf{Track 3: Naturalistic Driving Action Recognition.}
Based on 10 hours of videos collected from 10 diverse drivers, each team was asked to submit one text file containing the details of one identified activity on each line. The details include the start and end times of the activity and corresponding video file information. Table~\ref{tab:driver:view} shows the three types of in-vehicle camera views, and Figure~\ref{fig:track3:camera} shows the camera mounting setup. Although normal forward driving was listed as one of the distracting activities, it was not considered for evaluation. Teams' performance is measured by F-1 score, and the team with the highest F1 score becomes the winner of this track.

\begin{table}[t]
\caption{The three in-vehicle camera views for driver behavior recognition.}
\centering
\begin{tabular}{|c|c|}
\hline
Camera     & Location                \\ \hline
Dash Cam 1 & Dashboard               \\ \hline
Dash Cam 2 & Behind rear view mirror \\ \hline
Dash Cam 3 & Top right side window   \\ \hline
\end{tabular}
\label{tab:driver:view}
\end{table}

\textbf{Track 4: Multi-Class Product Recognition \& Counting for Automated Retail Checkout.} Teams were requested to perform retail object recognition and subsequently counting for automatic retail checkout. Given the test scenario of a retail staff moving retail objects across the area of interest, participant teams should report the object ID as well as the timestamp it appears. For the first time in AI City Challenge, we provide only synthetic data for model training, where the synthetic dataset is created using the 3D scans of retail objects.

%%%%%%%%%%%%%%%%%%%%%%%%%%%%%%%%%%%%%%%%%%%%%%%%%%%%%%%%%%
%%%%%%%%%%%%%%%%%%%%%%%%%%%%%%%%%%%%%%%%%%%%%%%%%%%%%%%%%%
\section{Datasets}
\label{sec:dataset}

For Track 1 and Track 2, the data were collected from traffic cameras placed in multiple intersections of a mid-size U.S. city. The homography matrices for mapping the ground plane to the image plane are provided. The privacy issue has been addressed by redacting vehicle license plates and human faces. The manually annotated NL descriptions are provided in the task of Track 2. As for Track 3, the participating teams are presented with synthetic naturalistic data of the driver collected from three camera locations inside the vehicle (while the driver is pretending to be driving). In Track 4, participants identify/classify products when a customer is hand holding items in front of the checkout counter. The products may be visually very similar or occluded by hands and other objects. Synthetic images are provided for training, while evaluations are conducted on real test videos. 

Specifically, we have provided the following datasets for the challenge this year: 
(1) \textit{CityFlowV2}~\cite{Tang19CityFlow, Naphade19AIC19, Naphade20AIC20, Naphade21AIC21} for Track 1 on MTMC tracking, (2) \textit{CityFlow-NL}~\cite{feng2021cityflownl} for Track 2 on NL based vehicle retrieval, (3) \textit{SynDD1} for Track 3 on naturalistic driving action recognition, and (4) The \textit{Automated Retail Checkout (ARC)} dataset for Track 4 on multi-class product counting and recognition.

%%%%%%%%%%%%%%%%%%%%%%%%%%%%%%%%%%%%%%%%%%%%%%
\subsection{The {\bf \textit{CityFlowV2}} Dataset}

We first introduced the \textit{CityFlow} benchmark~\cite{Tang19CityFlow} in the 3rd AI City Challenge~\cite{Naphade19AIC19}. To our knowledge, \textit{CityFlow} was the first city-scale benchmark for MTMC vehicle tracking. In 2021, we have upgraded the dataset by refining the annotations and introducing a new test set referred to as \textit{CityFlowV2}. The validation set of \textit{CityFlowV2} is the same as the original test set of \textit{CityFlow}. 

\textit{CityFlowV2} contains 3.58 hours (215.03 minutes) of videos collected from $46$ cameras spanning 16 intersections. The distance between the two furthest simultaneous cameras is $4$ km. The dataset covers a diverse set of location types, including intersections, stretches of roadways, and highways. The dataset is divided into six scenarios. Three of the scenarios are used for training, two are for validation, and the remaining scenario is for testing. In total, the dataset contains $313,931$ bounding boxes for $880$ distinct annotated vehicle identities. Only vehicles passing through at least two cameras have been annotated. The resolution of each video is at least 960p and the majority of the videos have a frame rate of 10 frames per second. Additionally, in each scenario, the offset from the start time is available for each video, which can be used for synchronization. 

The \textit{VehicleX} dataset~\cite{Yao19VehicleX, Tang19PAMTRI} was also made available to the teams, which contains a large number of different types of backbone models and textures for 3D vehicle synthesis. Rendered by Unity~\cite{juliani2018unity}, a team can potentially generate an unlimited number of identities and images by editing various attributes, including orientations, camera parameters, and lighting settings. With these attributes, participants can perform multi-task learning, which can potentially improve the accuracy of re-identification (ReID)~\cite{Tang19PAMTRI, lin2019improving}. 

\subsection{The {\bf \textit{CityFlow-NL}} Dataset}

The \textit{CityFlow-NL} benchmark~\cite{feng2021cityflownl} consists of $666$ target vehicles in $3,598$ single-view tracks from $46$ calibrated cameras and $6,784$ unique NL descriptions. For each target, NL descriptions were provided by at least three crowd-sourcing workers, to better capture realistic variations and ambiguities that are expected in the real-world application domains. The NL descriptions provide information of the vehicle color, vehicle maneuver, traffic scene, and relations with other vehicles. 

For the tracked-vehicle retrieval by NL task, we utilized the \textit{CityFlow-NL} benchmark in a \textit{single-view} setup. For each single-view vehicle track, we bundled it with a query consisting of three different NL descriptions for training. During evaluation, the goal is to retrieve and rank vehicle tracks based on the given NL queries. This variation of the proposed \textit{CityFlow-NL} contains $2,155$ tracks of vehicles with three unique NL descriptions each. Additionally, $184$ unique vehicle tracks together with $184$ query sets (each annotated with three NL descriptions) are gathered and organized for testing.

%%%%%%%%%%%%%%%%%%%%%%%%%%%%%%%%%%%%%%%%%%%%%%
\subsection{The {\bf \textit{SynDD1}} Dataset}

\textit{SynDD1} \cite{https://doi.org/10.48550/arxiv.2204.08096} consists of 30 video clips in the training set and 30 videos in the test set. The data were collected using three in-vehicle cameras positioned at locations:  on the dashboard, near the rear-view mirror, and on the top right-side window corner as shown in Table~\ref{tab:driver:view} and Figure~\ref{fig:track3:camera}. The videos were recorded at 30 frames per second at a resolution of 1920×1080 and were manually synchronized for the three camera views. Each video is approximately 10 minutes in length and contains all 18 distracted activities shown in Table~\ref{tab:driving:activities}. These enacted activities were executed by the driver with or without an appearance block such as a hat or sunglasses in random order for a random duration. There were six videos for each driver: three videos in sync with an appearance block and three other videos in sync without any appearance block.

\begin{table}[]
\caption{The list of distracted driving activities in the {\it SynDD1} dataset.}
\centering
\begin{tabular}{|c|c|}
\hline
Sr. no. & Distracted driver behavior      \\ \hline
1       & Normal forward driving            \\ \hline
2       & Drinking                          \\ \hline
3       & Phone call (right)                \\ \hline
4       & Phone call (left)                 \\ \hline
5       & Eating                            \\ \hline
6       & Texting (right)                      \\ \hline
7       & Texting (left)                       \\ \hline
8       & Hair / makeup                     \\ \hline
9       & Reaching behind                   \\ \hline
10      & Adjusting control panel           \\ \hline
11      & Picking up from floor (driver)    \\ \hline
12      & Picking up from floor (passenger) \\ \hline
13      & Talking to passenger at the right \\ \hline
14      & Talking to passenger at backseat  \\ \hline
15      & Yawning                           \\ \hline
16      & Hand on head                      \\ \hline
17      & Singing with music                \\ \hline
18      & Shaking or dancing with music     \\ \hline
\end{tabular}
\label{tab:driving:activities}
\end{table}

%%%%%%%%%%%%%%%%%%%%%%%%%%%%%%%%%%%%%%%%%%%%%%

\begin{figure}[t]
\centering
\includegraphics[width=0.47\textwidth]{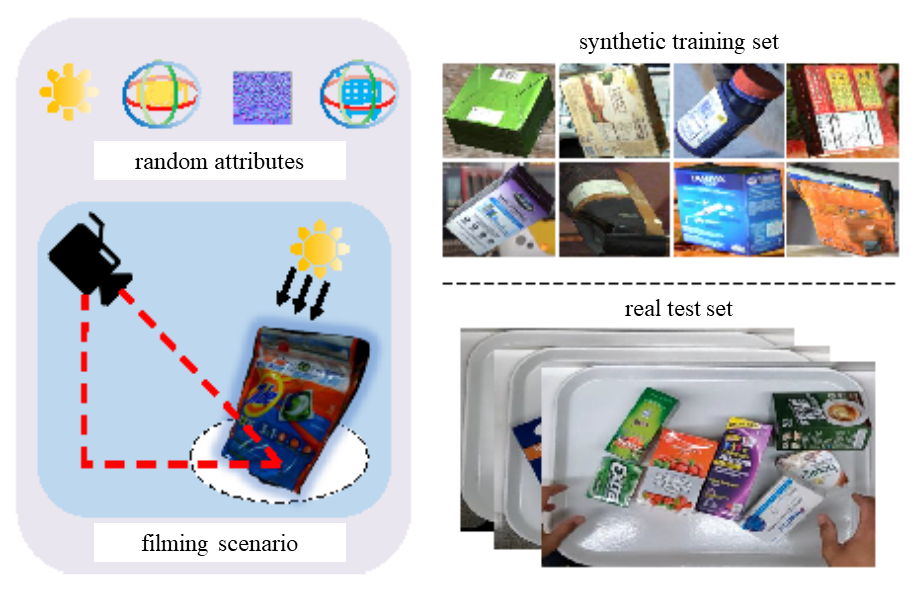}
\caption{The {\it Automated Retail Checkout (ARC)} dataset includes two parts: synthetic data for model training and real-world data for model validation and testing.}
\label{img:retail_datasets}
\end{figure}

\subsection{The {\bf \textit{Automated Retail Checkout (ARC)}} Dataset}

The {\it Automated Retail Checkout (ARC)} dataset includes two parts: synthetic data for model training and real data for model validation and testing. 

The synthetic data for Track 4 is created using the pipeline from~\cite{yao2022attribute}. Specifically, we collected 116 scans of real-world retail objects obtained from supermarkets in 3D models. Objects class ranges from daily necessities, food, toys, furniture, household, \etc. A total of $116,500$ synthetic images were generated from these $116$ 3D models. Images were filmed with a scenario as shown in Figure~\ref{img:retail_datasets}. Random attributes including random object placement, camera pose, lighting, and backgrounds were adopted to increase the dataset diversity. Background images were chosen from Microsoft COCO~\cite{lin2014microsoft}, which has diverse scenes suitable for serving as natural image backgrounds. 

In our test scenario, the camera was mounted above the checkout counter and facing straight down, while a customer was enacting a checkout action by ``scanning'' objects in front of the counter in a natural manner. Several different customers participated, where each of them scanned slightly differently. There was a shopping tray placed under the camera to indicate where the AI model should focus. In summary, we obtained approximately $22$ minutes of video, and the videos were further split into \textit{testA} and \textit{testB} sets. The former amounts to $20\%$ of recorded test videos that were used for model validation and inference code development. The latter accounts for $80\%$ of the videos, which were reserved for testing and determining the ranking of participant teams.

%%%%%%%%%%%%%%%%%%%%%%%%%%%%%%%%%%%%%%%%%%%%%%%%%%%%%%%%%%
%%%%%%%%%%%%%%%%%%%%%%%%%%%%%%%%%%%%%%%%%%%%%%%%%%%%%%%%%%
\section{Evaluation Methodology}
\label{sec:eval}

Similar to previous AI City Challenges~\cite{Naphade18AIC18,Naphade19AIC19,Naphade20AIC20,Naphade21AIC21}, teams submitted multiple runs to an \textbf{online evaluation system} that automatically measured the effectiveness of results from the submissions. Team submissions were limited to five per day and a total of twenty submissions per track. Any submissions that led to a format or evaluation error did not count against a team's daily or maximum submission totals. During the competition, the evaluation system showed the team's own performance, along with the top-3 best scores on the leader boards, without revealing the identities of those teams. To discourage excessive fine-tuning to improve performance, the results shown to the teams prior to the end of the challenge were computed on a 50\% subset of the test set for each track. After the challenge submission deadline, the evaluation system revealed the full leader boards with scores computed on the entire test set for each track.

Teams competing for the challenge prizes were not allowed to use external data or manual labeling to fine-tune the performance of their model, and those results were published on the {\bf Public} leader board. Teams using additional external data or manual labeling were allowed to submit to a separate {\bf General} leader board.

%%%%%%%%%%%%%%%%%%%%%%%%%%%%%%%%%%%%%%%%%%%%%%%%%%%%%%%%%%%
\subsection{Track 1 Evaluation}

The Track 1 task was evaluated based on the IDF1 score~\cite{Ristani16Performance} similar to the evaluation of Track 3 of our 2021 Challenge~\cite{Naphade21AIC21}). The IDF1 score measures the ratio of correctly identified detections over the average number of ground truth and computed detections. The evaluation tool provided with our dataset also computed other evaluation measures adopted by the \textit{MOTChallenge}~\cite{Bernardin2008,Li09Learning}. These provided measures include the multiple object tracking accuracy (MOTA), multiple object tracking precision (MOTP), mostly tracked targets (MT), and false alarm rate (FAR). However, these measures were not used for ranking purposes in our contest. The measures that were displayed in the evaluation system were IDF1, IDP, IDR, Precision (detection), and Recall (detection).

%%%%%%%%%%%%%%%%%%%%%%%%%%%%%%%%%%%%%%%%%%%%%%%%%%%%%%%%%%%
\subsection{Track 2 Evaluation} 
\label{sec:track2:eval}

Track 2 was originally inaugurated as Track 5 of our 2021 Challenge~\cite{Naphade21AIC21}. The evaluation was performed using standard metrics for retrieval tasks~\cite{manning2008introduction}, namely the Mean Reciprocal Rank as the evaluation metric. In addition, Recall@5, Recall@10, and Recall@25 were also evaluated for all models but were not used in the ranking. For a given set $Q$ of queries, the MRR score is computed as
\begin{equation}
    \text{MRR}={\frac{1}{|Q|}}\sum _{i=1}^{|Q|}{\frac {1}{{\text{rank}}_{i}}},
\end{equation}
where ${\text{rank}}_{i}$ refers to the ranking position of the first relevant document for the $i$-th query, and $|Q|$ is the set size. 

%%%%%%%%%%%%%%%%%%%%%%%%%%%%%%%%%%%%%%%%%%%%%%%%%%%%%%%%%%%
\subsection{Track 3 Evaluation}
\label{sec:track1:eval}

Evaluation for Track 3 was based on model activity identification performance, measured by the standard F1-score metric. For the purpose of computing the F1-score, a true-positive (TP) activity identification was considered when an activity was correctly identified (matching activity ID) as starting within one second of the start time and ending within one second of the end time of the activity. Only one activity was allowed to match to any ground truth activities. Any other reported activities that were not TP activities were marked as false-positive (FP). Finally, ground truth activities that were not correctly identified were marked as false-negative (FN). 

%%%%%%%%%%%%%%%%%%%%%%%%%%%%%%%%%%%%%%%%%%%%%%%%%%%%%%%%%%%
\subsection{Track 4 Evaluation}
\label{sec:track4:eval}

Evaluation for Track 4 was also based on model identification performance, measured by the F1-score metric. For the purpose of computing the F1-score, a true-positive (TP) identification was considered when an object was correctly identified within the region of interest, {\em i.e.}, the object class was correctly determined, and the object was identified within the time that the object was over the white tray. Only one object was allowed to match to any ground truth object. A false-positive (FP) was an identified object that was not a TP identification. Finally, a false-negative (FN) identification was a ground truth object that was not correctly identified.

%%%%%%%%%%%%%%%%%%%%%%%%%%%%%%%%%%%%%%%%%%%%%%%%%%%%%%%%%%
%%%%%%%%%%%%%%%%%%%%%%%%%%%%%%%%%%%%%%%%%%%%%%%%%%%%%%%%%%
\section{Challenge Results}
\label{sec:results}

Tables~\ref{table:1},~\ref{table:2},~\ref{table:3}, and ~\ref{table:4} summarize the leader boards for Track 1 (city-scale MTMC vehicle tracking), Track 2 (NL based vehicle retrieval), Track 3 (natural driving action recognition), and Track 4 (multi-class product counting and recognition), respectively.

%%%%%%%%%%%%%%%%%%%%%%%%%%%%%%%%%%%%%%%%%%%
\subsection{Summary for the Track 1 Challenge}

\begin{table}[t]
\caption{Summary of the Track 1 leader board.}
\label{table:1}
\centering
\footnotesize
\begin{tabular}{|c|c|c|c|}
\hline
Rank & Team ID & Team and paper & Score \\
\hline\hline
1 & 28 & Baidu~\cite{Baidu22MTMCT} & {\bf 0.8486} \\
\hline
2 & 59 & BOE~\cite{BOE22MTMCT} & 0.8437 \\
\hline
3 & 37 & Alibaba~\cite{Alibaba22MTMCT} & 0.8371 \\
\hline
4 & 50 & Fraunhofer IOSB~\cite{Fraunhofer22MTMCT} & 0.8348 \\
\hline
10 & 94 & SKKU~\cite{SKKU22MTMCT} & 0.8129 \\
\hline
18 & 4 & HCMIU~\cite{HCMIU22MTMCT} & 0.7255 \\
\hline
10 (General) & 107 & SUTPC~\cite{SUTPC22MTMCT} & 0.8285 \\
\hline
\end{tabular}
\vspace{-0.4cm}
\end{table}

Most teams applied the typical workflow of MTMC tracking which includes four steps. (1) The first step is vehicle detection. The best performing teams utilized the state-of-the-art detectors such as YOLOv5~\cite{glenn_jocher_2020_4154370} and Cascade R-CNN~\cite{Cai18CascadeRCNN}. (2) Secondly, teams exploited ReID models to extract robust appearance features. Some of them~\cite{Baidu22MTMCT, Alibaba22MTMCT} concatenated the feature vectors from multiple models for enhancing the descriptors. The HCMIU team~\cite{HCMIU22MTMCT} leveraged synthetic data and re-ranking with contextual constraints for domain adaptation and generated reliable feature embeddings. (3) Single-camera tracklets were formed based on the detection results (bounding boxes) and the corresponding feature embeddings. The top-ranked team from Baidu~\cite{Baidu22MTMCT} employed DeepSORT~\cite{Wojke17} for single-camera tracking. The BOE team~\cite{BOE22MTMCT} with 2nd rank incorporated augmented tracks prediction using MedianFlow, multi-level association, and zone-based merging to optimize the tracklets. The team from Fraunhofer IOSB~\cite{Fraunhofer22MTMCT} further enhanced single-camera tracklets by appearance-based tracklet splitting, clustering, and track completion. The SUTPC team~\cite{SUTPC22MTMCT} proposed an occlusion-aware module to connect broken tracklets. (4) The most important component for MTMC tracking is inter-camera association. Most teams built similarity matrices with appearance and spatio-temporal information and applied hierarchical clustering. For example, the team from Baidu~\cite{Baidu22MTMCT} used \textit{k}-reciprocal nearest neighbors for clustering with constraints of traveling time, road structures, and traffic rules to reduce searching space. Likewise, the Alibaba team~\cite{Alibaba22MTMCT} introduced a zone-gate and time-decay based matching mechanism. 

%%%%%%%%%%%%%%%%%%%%%%%%%%%%%%%%%%%%%%%%%%%
\subsection{Summary for the Track 2 Challenge}

\begin{table}[t]
  \caption{Summary of the Track 2 leader board.}
  \label{table:2}
  \centering
  \footnotesize
  \begin{tabular}{|c|c|c|c|}
    \hline
    Rank & Team ID & Team and paper & Score (MRR) \\
    \hline\hline
    1 & 176 & Baidu-SYSU~\cite{BaiduSYSU22NLRetrieval} & {\bf 0.6606} \\
    \hline
    3 & 4 & HCMIU~\cite{HCMIU22NLRetrieval} & 0.4773 \\
    \hline
    4 & 183 & Megvii~\cite{Megvii22NLRetrieval} & 0.4392 \\
    \hline
    5 & 91 & HCMUS-UDayton~\cite{HCMUSUDayton22NLRetrieval} & 0.3611 \\
    \hline
    7 & 10 & Terminus-CQUPT~\cite{TerminusCQUPT22NLRetrieval} & 0.3320 \\
    \hline
    9 & 24 & BUPT-ChinaMobile~\cite{BUPTChinaMobile22NLRetrieval} & 0.3012 \\
    \hline
  \end{tabular}
  \vspace{-0.4cm}
\end{table}

For the task of tracked-vehicle retrieval by NL descriptions, all teams used ReID inspired approaches to measure the similarities between the visual features (both local and global) and the language query features. InfoNCE losses were used by all participating teams to train for the text-to-image retrieval task. Additionally, to represent the NL descriptions, all participating teams utilized some forms of pre-trained sentence embedding model, \eg BERT~\cite{devlin2018bert}. The team of \cite{BaiduSYSU22NLRetrieval} used an NL parser to obtain the color, type, and motion of tracked-vehicles. These attributes were used in addition to the ReID-based approach to post-process the retrieval results. Vehicle motion is an essential part of the NL descriptions in \textit{CityFlow-NL}. Therefore, some teams~\cite{BUPTChinaMobile22NLRetrieval,TerminusCQUPT22NLRetrieval,Megvii22NLRetrieval} used the global motion image introduced by Bai \etal~\cite{AlibabaUTSZJU21NLRetrieval} to construct a stream for vehicle motion. The Megvii team \cite{Megvii22NLRetrieval} introduced an improved motion image based on the inter-frame IoU of the tracked targets.

The best performing team~\cite{Baidu22MTMCT} presented a state-of-the-art tracked-vehicle retrieval by NL system by training a cosine similarity between language query features and visual features. A \textit{Target Vehicle Attribute Enhancement} module post-processed and re-weighted the retrieval results based on the parsed language attributes. This module improved the test performance from 40.73\% to 56.52\%. The team of \cite{HCMIU22NLRetrieval} proposed a \textit{Semi-supervised Domain Adaptation} training process and performed motion analysis and post-processing with pruning of retrieval results. In addition to the improved motion image, the Megvii team \cite{Megvii22NLRetrieval} proposed hard test samples mining and short-distance relationship mining to distinguish visually similar vehicles and the relations between them. The team of \cite{HCMUSUDayton22NLRetrieval} implemented a post-processing step to refine the retrieval results specifically for the straight-following case. Local instance and motion features, the motion image, and video clip embeddings were used to build a quad-stream retrieval model in \cite{TerminusCQUPT22NLRetrieval}. Lastly, the team of \cite{BUPTChinaMobile22NLRetrieval} proposed a multi-granularity loss function, which is a pair-wise InfoNCE loss between NL streams and visual streams, to formulate the ReID problem.

%%%%%%%%%%%%%%%%%%%%%%%%%%%%%%%%%%%%%%%%%%%
\subsection{Summary for the Track 3 Challenge}

\begin{table}[t]
\caption{Summary of the Track 3 leader board.}
\label{table:3}
\centering
\footnotesize
\begin{tabular}{|c|c|c|c|}
\hline
Rank & Team ID & Team and paper & Score \\
\hline\hline
1 & 72 & Viettel~\cite{Viettel22ActionRecognition} & {\bf 0.3492} \\
\hline
2 & 43 & Tencent-THU~\cite{TencentTHU22ActionRecognition} & 0.3295 \\
\hline
3 & 97 & CyberCore~\cite{CyberCore22ActionRecognition} & 0.3248 \\
\hline
4 & 15 & Oppo-ZJU-ECUST~\cite{OppoZJUECUST22ActionRecognition} & 0.3154 \\
\hline
5 & 78 & USF~\cite{USF22ActionRecognition} & 0.2921 \\
\hline
6 & 16 & BUPT~\cite{BUPT22ActionRecognition} & 0.2905 \\
\hline
7 & 106 & WHU~\cite{WHU22ActionRecognition} & 0.2902 \\
\hline
9 & 54 & TUE~\cite{TUE22ActionRecognition} & 0.2710 \\
\hline
10 & 95 & Tahakom~\cite{Tahakom22ActionRecognition} & 0.2706 \\
\hline
11 & 1 & SCU~\cite{SCU22ActionRecognition} & 0.2558 \\
\hline
\end{tabular}
\vspace{-0.4cm}
\end{table}

The methodologies of the top performing teams in Track 3 of the Challenge were based on the basic idea of activity recognition which involved: (1) classification of various distracted activities such as eating, texting, yawning, {\em etc.}, and (2) Temporal Action Localization (TAL) which determines the start and end time for each activity.  The best performing team, Viettel~\cite{Viettel22ActionRecognition}, utilized the 3D action recognition model X3D \cite{feichtenhofer2020x3d} to extract short temporal and spatial correlation together with a multi-view ensemble technique to classify the activity type. Post-processing was performed for localizing long temporal correlation to predict TAL. Their best score was 0.3492. The runner-up, Tencent-THU~\cite{TencentTHU22ActionRecognition} used the multi-scale vision transformer network for action recognition and sliding window classification for TAL. The third-place team, CyberCore~\cite{CyberCore22ActionRecognition} implemented the prediction of temporal location and classification simultaneously. The ConvNext~\cite{liu2022convnet} was used as backbone model for recognition. They applied two techniques: {\em learning without forgetting} and semi-weak supervised learning to avoid over-fitting and improve model performance.

%%%%%%%%%%%%%%%%%%%%%%%%%%%%%%%%%%%%%%%%%%%
\subsection{Summary for the Track 4 Challenge}

\begin{table}[t]
\caption{Summary of the Track 4 leader board.}
\label{table:4}
\centering
\footnotesize
\begin{tabular}{|c|c|c|c|}
\hline
Rank & Team ID & Team and paper & Score \\
\hline\hline
1 & 16 & BUPT~\cite{BUPT22AutomatedCheckout} & {\bf 1.0000} \\
\hline
2 & 94 & SKKU~\cite{SKKU22AutomatedCheckout} & 0.4783 \\
\hline
3 & 104 & SUST-Giga-ConcordiaU-NSU~\cite{SUSTGigaConcordiaUNSU22AutomatedCheckout} & 0.4545 \\
\hline
4 & 165 & Mizzou~\cite{Mizzou22AutomatedCheckout} & 0.4400 \\
\hline
7 & 117 & BUT~\cite{BUT22AutomatedCheckout} & 0.4167 \\
\hline
\end{tabular}
\end{table}

Most teams handled the task of auto retail checkout following the detection-tracking-counting (DTC) framework. (1) First, object detection is used to estimate the bounding boxes for retail objects. The best performing method~\cite{BUPT22AutomatedCheckout} used DetectoRS~\cite{qiao2021detectors} while other teams also used comparable detectors such as YOLOv5~\cite{glenn_jocher_2020_4154370} and Scaled-YOLOv4~\cite{wang2021scaled}. In order to obtain accurate object boundary, some teams further used segmentation to filter out occlusions such as the palms or other retail objects~\cite{BUPT22AutomatedCheckout,SUSTGigaConcordiaUNSU22AutomatedCheckout,BUT22AutomatedCheckout}. For example, the BUT team masked off the human body regions using image inpainting~\cite{BUT22AutomatedCheckout}. (2) Second, based on the detection results, single-camera tracking is performed to get the tracklets. The top-ranked team employed DeepSORT~\cite{Wojke17} for single-camera tracking~\cite{BUPT22AutomatedCheckout,SKKU22AutomatedCheckout,Mizzou22AutomatedCheckout}. And some others used association methods like ByteTrack~\cite{zhang2021bytetrack}. Notably, to bridge the large domain gaps between the synthetic training set and real-world test set, various transformations were applied to the training set. Many teams used real-world background images when training the detection and segmentation networks~\cite{BUPT22AutomatedCheckout,BUT22AutomatedCheckout,SKKU22AutomatedCheckout}. (3) With the single-camera tracklets, post-processing is applied to get the timestamp (\ie, counting) when the object is in the area of interest. For example, the BUPT team~\cite{BUPT22AutomatedCheckout} proposed an algorithm to link the potential broken tracklets.

%%%%%%%%%%%%%%%%%%%%%%%%%%%%%%%%%%%%%%%%%%%%%%%%%%%%%%%%%%
%%%%%%%%%%%%%%%%%%%%%%%%%%%%%%%%%%%%%%%%%%%%%%%%%%%%%%%%%%
\section{Discussion and Conclusion}
\label{sec:conclusion}

The 6th AI City Challenge continues to attract worldwide research community participation in terms of both quantity and quality. We provide a few observations below.

In Track 1, teams continue to push the state-of-the-art on the {\it CityFlow} benchmark by introducing new mechanisms to refine the single-camera tracklets and improve the hierarchical clustering of inter-camera association. Some of the teams exploited the synthetic data and utilized domain adaptation to enhance the ReID features. However, most of the proposed methods had to rely on prior knowledge of the scene and manual definition of entry/exit zones, which may not be feasible for a real-world system where there are thousands of cameras. The scene information will need to be extracted automatically from the open geographic data based on the GPS coordinates. Moreover, due to the short duration of the test set, all the proposed methods are based on batch processing. Those methods are not ready to be scaled up for live streaming applications in real world. 

In Track 2, we updated the {\it CityFlow-NL} benchmark with new language annotations and training/test splits. Teams were challenged to apply knowledge across computer vision and NLP to the retrieval task of tracked-vehicles using a natural language query. Participant teams built retrieval systems based on the findings from the previous AI City Challenge. Various approaches based on ReID approaches were introduced by teams to learn representative motion and visual appearance features. Post-processing of retrieval results based on the keywords of relations and motions in the NL descriptions were introduced by participating teams to further improve the retrieval results. In Track 2, with the newly curated train/test splits, we have seen major improvements on the retrieval performance of the top-ranked teams to achieve a Recall @ 5 (out of 185) over 70\%. However, a performance gap between best performing models still exists. Finally, how to best post-process and prune based on the keyword extractions from the NL queries remains the main difficulty.

In Track 3, participant teams worked on the {\it SynDD1}~\cite{https://doi.org/10.48550/arxiv.2204.08096} benchmark and considered it as a Driver Activity Recognition problem with the aim to design an efficient detection method to identify a wide range of distracted activities. This challenge addressed two problems, classification of driver activity as well as temporal localization to identify their start and end time. To this end, participant teams have spent significant efforts in optimizing algorithms as well as implementing the pipelines for performance improvement. They tackled the problem by adopting techniques including the vision transformers \cite{BUPT22ActionRecognition,CyberCore22ActionRecognition,OppoZJUECUST22ActionRecognition,TencentTHU22ActionRecognition} and action classifiers \cite{Tahakom22ActionRecognition,SCU22ActionRecognition,WHU22ActionRecognition,TUE22ActionRecognition,Viettel22ActionRecognition}. Both activity recognition and temporal action localization are still open research problems that require more in-depth study. More clean data and ground truth labels can clearly improve the development and evaluation of the research progress. We plan to increase the size and quality of the {\it SynDD1} dataset, with a hope that it will significantly boost future research in this regard. 

The main thrust of Track 4 this year was the evaluation of retail object recognition and counting methods on the edge IoT devices. To this end, significant efforts have been made by participant teams in implementing pipelines as well as optimizing algorithms for performance improvement. Among top-performing teams, the detection-tracking-counting (DTC) framework remained the most popular scheme~\cite{BUPT22AutomatedCheckout,SKKU22AutomatedCheckout,Mizzou22AutomatedCheckout,BUT22AutomatedCheckout}. Within the DTC framework, object tracking as well as the segmentation were the focus. Notably, the domain gap between synthetic training and real testing data remains the main difficulty for the implementation of the DTC framework, as they have large difference on filming scenarios. Many teams utilized various image transformations to reduce such gaps, and this led to significant improvement on accuracy~\cite{BUPT22AutomatedCheckout,BUT22AutomatedCheckout,SKKU22AutomatedCheckout}.   

{\bf Future work.} We envision that the future editions of the AI City Challenge will continue to push the boundary of advancing the state-of-the-art and bridging the gap between experimental methods and their real-world deployment to make environments around us smarter. With this edition we have expanded the breadth of the challenge to cover multiple verticals including transportation and retail sectors. We hope to enrich the challenge tracks with larger data sets going forward. We also hope to add new tasks that push the state of the art in other aspects of AI Cities. 

%%%%%%%%%%%%%%%%%%%%%%%%%%%%%%%%%%%%%%%%%%%%%%%%%%%%%%%%%%
%%%%%%%%%%%%%%%%%%%%%%%%%%%%%%%%%%%%%%%%%%%%%%%%%%%%%%%%%%
\section{Acknowledgment}

The datasets of the 6th AI City Challenge would not have been possible without significant contributions from the Iowa DOT and an urban traffic agency in the United States. This Challenge was also made possible by significant data curation help from the NVIDIA Corporation and academic partners at the Iowa State University, Boston University, and Australian National University. We would like to specially thank Paul Hendricks and Arman Toorians from the NVIDIA Corporation for their help with the retail dataset.

%%%%%%%%%%%%%%%%%%%%%%%%%%%%%%%%%%%%%%%%%%%%%%%%%%%%%%%%%%
%%%%%%%%%%%%%%%%%%%%%%%%%%%%%%%%%%%%%%%%%%%%%%%%%%%%%%%%%%
%\newpage

{\small
\bibliographystyle{ieee_fullname}
%\bibliography{aicity22, aicity21, aicity20, aicity19, aicity18, aicity17}
\bibliography{main}
}

\end{document}